\documentclass[journal]{IEEEtran}

\usepackage{bm}
\usepackage{array}
\usepackage{threeparttable}
\usepackage{multirow}
\usepackage{booktabs}
\usepackage{graphics}
\usepackage[caption=false,font=footnotesize]{subfig}
\usepackage{epstopdf}
\usepackage{epsfig} 
\usepackage{algorithm}
\usepackage{algorithmic}
\usepackage{enumerate}
\usepackage{setspace}
\usepackage{cite}
\usepackage{amsmath}
\usepackage{amsfonts}
\usepackage{color}

\begin{document}

\title{A Roadmap-Path Reshaping Algorithm for Real-Time Motion Planning}

\author{Chaoyi~Sun,
		Qing~Li,
		and~Li~Li,~\IEEEmembership{Fellow,~IEEE}

\thanks{Manuscript received on February 28th, 2019. This work was supported in part by Beijing Municipal Science and Technology Program under Grant D15110900280000. (Corresponding author: \emph{Li Li})}
\thanks{C. Sun and Q. Li are with the Department of Automation, Tsinghua University, Beijing China, 100084.}
\thanks{L. Li are with the Department of Automation, BNRist, Tsinghua University, Beijing, China, 100084 (e-mail: li-li@tsinghua.edu.cn).}}

\markboth{}%
{Shell \MakeLowercase{\textit{Sun et al.}}: A Roadmap-Path Reshaping Algorithm for Real-Time Motion Planning}

\maketitle

\begin{abstract}
Real-time motion planning is a vital function of robotic systems. Different from existing roadmap algorithms which first determine the free space and then determine the collision-free path, researchers recently proposed several convex relaxation based smoothing algorithms which first select an initial path to link the starting configuration and the goal configuration and then reshape this path to meet other requirements (e.g., collision-free conditions) by using convex relaxation. However, convex relaxation based smoothing algorithms often fail to give a satisfactory path, since the initial paths are selected randomly. Moreover, the curvature constraints were not considered in the existing convex relaxation based smoothing algorithms. In this paper, we show that we can first grid the whole configuration space to pick a candidate path and reshape this shortest path to meet our goal. This new algorithm inherits the merits of the roadmap algorithms and the convex feasible set algorithm. We further discuss how to meet the curvature constraints by using both the Beamlet algorithm to select a better initial path and an iterative optimization algorithm to adjust the curvature of the path. Theoretical analyzing and numerical testing results show that it can almost surely find a feasible path and use much less time than the recently proposed convex feasible set algorithm.
\end{abstract}


\IEEEpeerreviewmaketitle

\section{Introduction}
\label{section1}
\IEEEPARstart{M}{otion} planning refers to the problem of finding a collision-free and dynamically-feasible path between the pre-specified starting configuration and the goal configuration in obstacle-cluttered environments \cite{ref1}, \cite{ref2}. Since it is often hard to achieve both collision-free and smooth paths simultaneously, we usually decompose the problem into two stages: first finding a collision-free path and then smoothing it \cite{ref3}.

There were various roadmap algorithms \cite{ref4} -\cite{ref9} to compute collision-free paths. These algorithms first determine the free space and then find the collision-free paths characterized by a few waypoints within the free space. However, it is difficult for roadmap algorithms to simultaneously satisfy both dynamic constraints and exploring the configuration space. Their returned paths tend to be unsmooth and may have sharp turns.

To make the path more suitable for mobile robots, researchers proposed different path smoothing algorithms \cite{ref10}-\cite{ref11} which can be divided into three kinds: interpolation-based, shortcut, and optimization-based algorithms \cite{ref3}, \cite{ref12}, \cite{ref13}.

Interpolation-based algorithms attempt to relocate waypoints to fit some types of curves (e.g., polynomial curve \cite{ref14}, spline \cite{ref14}, or NURBS curve \cite{ref15}) that have good smoothness. Shortcut algorithms \cite{ref16} check and replace the jerky portions of a path with some transition curves (e.g., Dubin’s curve \cite{ref17}, clothoid \cite{ref18}, or hypocycloid \cite{ref19}). However, both the two kinds of algorithms restrict the moving scopes of the waypoints and cannot significantly change the shape of the whole path.

In contrast, optimization-based algorithms allow waypoints to move significantly away from their original locations. For example, the Convex Elastic Smoothing (CES) algorithm \cite{ref3} uses a set of "bubbles" along the initial path to approximate the free space and let the waypoints to move within these "bubbles". Then, the CES algorithm solves two convex optimization problems alternately for path smoothing or speed optimization. Furthermore, the Convex Feasible Set (CFS) algorithm \cite{ref20} -\cite{ref23} uses the intersection of convex cones to define a larger free space and leads to a faster convergence speed. To distinguish optimization-based algorithms from interpolation-based and shortcut algorithms, we would like to call their iteration process as "path reshaping" rather than "path smoothing".

However, many optimization-based algorithms neglect the selection of the initial path and just choose a random path to start. As a result, these algorithms often fail to give a valid path, because the properties of the final path still highly depend on the initial path in optimization-based algorithms, although we have assigned the path more deformation capability.

To solve this problem, we propose a new algorithm to generate collision-free and smooth paths by combining the merits of the roadmap algorithm and the path reshaping algorithm. Specifically, we first grid the whole configuration space and dilate obstacles to find an appropriate initial path by Dijkstra’s algorithm \cite{ref24}. Then, we apply the CFS algorithm to reshape it. Gridding the configuration and dilating obstacles can guarantee the existence of a feasible final path and also help reduce the time cost of reshaping. To further accelerate our algorithm, we propose a modified algorithm by using the divide-and-conquer strategy.
We adopt the Beamlet algorithm \cite{ref25} to further select an initial path that is more suitable for the given curvature constraints. We also design an iterative optimization algorithm to adjust the path to rigorously meet the curvature constraints.

Numerical testing results show that our proposed algorithm can almost surely find a feasible path and requires less computation time, if compared to the CFS algorithm. To give a detailed explanation of our finding, the rest of this paper is organized as follows. Section \ref{section2} formulates the motion planning problem as an optimization problem and then introduces the CFS algorithm as a basis for further discussions. Section \ref{section3} presents our new algorithm, proves its feasibility, and explains its time complexity. Section \ref{section4} further discusses how to handle curvature constraints for the initial and final path. Section \ref{section5} presents some numerical testing results to validate our new algorithm. Finally, Section \ref{section6} concludes the paper.

\section{Problem Formulation}
\label{section2}
\subsection{The Optimiaztion Problem}
\label{subsection2_1}
Our goal is to find a path $\pmb{x}$ that is characterized by a series of $(n+1)$ waypoints, i.e. $\pmb{x}=[\pmb{x}_0^T,\pmb{x}_1^T,\dots,\pmb{x}_n^T]^T$, where $\pmb{x}_i\in\mathbb{R}^2$ represents the position of the robot at $i$th time in the configuration space.

In this paper, the robotic motion planning problem is formulated as the optimization problem \eqref{equ1}-\eqref{equ4} as below.
\begin{align}
\label{equ1}\max_{\pmb{x}}\quad &J(\pmb{x})=\pmb{x}(\pmb{V}^T\pmb{V}+\lambda\pmb{A}^T\pmb{A})\pmb{x}\\
\label{equ2}s.t.\quad & \pmb{x}_0=\pmb{x}_{start},~\pmb{x}_n=\pmb{x}_{end}\\
\label{equ3}&d(\pmb{x}_i,O_j)\ge d_{min},~i=0,\dots,n,~j=1,\dots,q\\
\label{equ4}&f_k(\pmb{x})\leq 0,~k=1,2,\dots,s
\end{align}
where $\pmb{x}_{start},\pmb{x}_{end}\in\mathbb{R}^2$ are the starting waypoint and the ending waypoint, respectively, and $O_1,O_2,\dots,O_q\in\mathbb{R}^2$ denote $q$ obstacles in the configuration space. The distances between the robot and the $j$th obstacle in the $i$th time is denoted as $d(\pmb{x}_i,O_j)$. $d_{min}$ is the minimum clearance between the robot and obstacles.

The objective function \eqref{equ1} is typically designed as a quadratic function, where the matrices $\pmb{V}\in\mathbb{R}^{2n\times 2(n+1)}$, $\pmb{A}\in\mathbb{R}^{2(n-1)\times 2(n+1)}$ are pre-defined weighting matrices \cite{ref20} that indicate the need to reduce the path length and acceleration/deceleration values of the path, respectively, and $\lambda\in\mathbb{R}^+$.

The major difficulty of the above optimization problem lies in its nonconvex constraints \eqref{equ3} that are designed for obstacle avoidance and its intricate constraints \eqref{equ4} on robots' dynamic. In this paper, we will first present how to handle the problem when constraints \eqref{equ3} alone are considered and then discuss how to solve the problem when both constraints \eqref{equ3}-\eqref{equ4} are considered.

\subsection{The Convex Feasible Set Algorithm}
\label{subsection2_2}
The CFS algorithm proposed in \cite{ref20} -\cite{ref23} assumes that every two consecutive waypoints are linked with line segments to form a piecewise linear path that connects the starting waypoint and the ending waypoint. Given the initial path, these waypoints will be adjusted in an iterative way so that the final path is feasible for the above constraints \eqref{equ2}-\eqref{equ3} and meanwhile minimize objective \eqref{equ1}. This setting is similar to the Convex Elastic Smoothing (CES) algorithm.

The difference lies in the setting of the local collision-free feasible regions in which the waypoints can be adjusted. The CES algorithm assumes that a sequence of "bubbles" is placed along the reference path to identify the local feasible region. In contrast, the CFS algorithm assumes the local feasible regions are the intersection of convex cones that can be found online. This new setting is less conservative and only add a few more computation costs.

The complete CFS algorithm can be summarized as Algorithm \ref{algorithm1} below. More details can be found in \cite{ref20}.

\begin{algorithm}
	\caption{The Convex Feasible Set Algorithm}
	\label{algorithm1}
	\begin{algorithmic}[1]
		\STATE\pmb{Initinalize}\\
		\STATE\quad Set an initial path $\pmb{x}^{(0)}=[\pmb{x}_0^{(0)T},\pmb{x}_1^{(0)T},\dots,\pmb{x}_n^{(0)T}]^T$.\\
		\STATE\quad Set the stopping threshold $\epsilon>0$.\\
		\STATE\quad Set $k=0$.\\
		\STATE\pmb{Loop}\\
		\STATE\quad Compute the convex feasible sets $F(\pmb{x}^{(k)})$.\\
		\STATE\quad\pmb{If} $F(\pmb{x}^{(k)})=\emptyset$\\
		\STATE\qquad Stop.\\
		\STATE\quad\pmb{Else}\\
		\STATE\qquad $\pmb{x}^{(k+1)}=\textrm{arg}\min\limits_{\pmb{x}\in F(\pmb{x}^{(k)})}J(\pmb{x})$.\\
		\STATE\quad\pmb{End If}\\
		\STATE\quad\pmb{If} $\lVert J(\pmb{x}^{(k+1)})-J(\pmb{x}^{(k)})\rVert<\epsilon$ or $\lVert\pmb{x}^{(k+1)}-\pmb{x}^{(k)}\rVert<\epsilon$\\
		\STATE\qquad Stop.\\
		\STATE\quad\pmb{End If}\\
		\STATE\quad $k=k+1$\\
		\STATE\pmb{End Loop}\\
		\STATE\pmb{Return} path $\pmb{x}^{(k+1)}=[\pmb{x}_0^{(k+1)T},\pmb{x}_1^{(k+1)T},\dots,\pmb{x}_n^{(k+1)T}]^T$.
	\end{algorithmic}
\end{algorithm}

However, testing results indicate that the CFS algorithm may fail to provide sufficient free spaces to reshape the path to avoid obstacles, if the initial path is inappropriately chosen, as shown in examples of Section \ref{subsection5_1}. Existing literature \cite{ref20}-\cite{ref23} neither discussed how many waypoints are needed for a particular problem, nor provided a way to set the initial path $\pmb{x}^{(0)}=[\pmb{x}_0^{(0)T},\pmb{x}_1^{(0)T},\dots,\pmb{x}_n^{(0)T}]^T$ to guarantee that we could find a feasible path finally.

\section{The Roadmap-Path Reshaping Algorithm}
\label{section3}
\subsection{The Basic Idea}
\label{subsection3_1}
\begin{figure}[htp]
	\centering
	\includegraphics[width=2.5in]{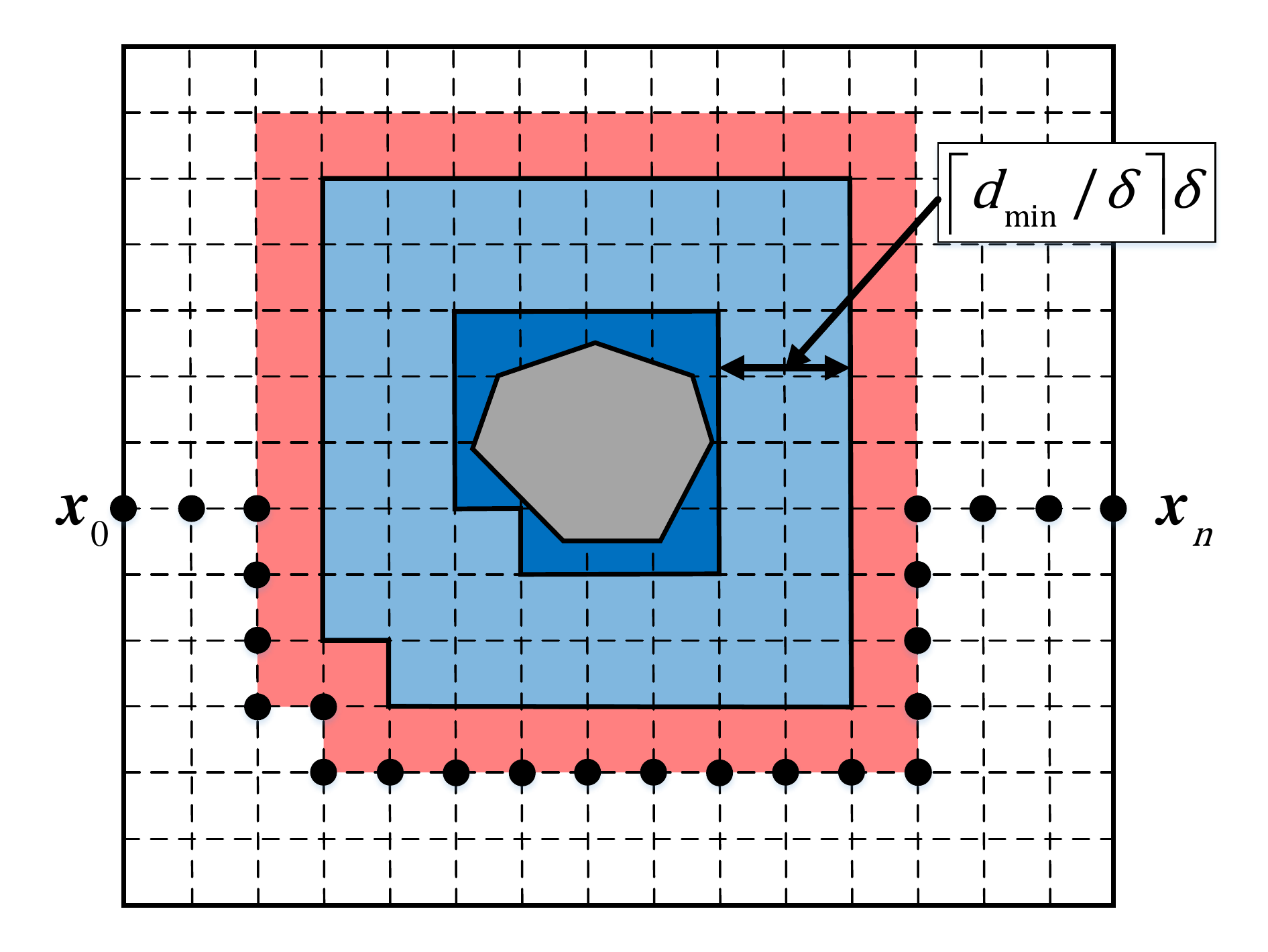}
	\caption{An illustration of the gridded configuration space and the dilation of obstacles.}
	\label{fig1}
\end{figure}

Inspired by the roadmap algorithms for path planning, we propose a new algorithm that can be roughly described as follows:

\underline{First}, uniformly grid the whole configuration space with a resolution level $\delta$ that is sufficiently small to guarantee the existence of a feasible path. Each intersection point of two gridlines will be taken as a node of the roadmap graph. Each line segment between two neighboring nodes will be taken as an arc of the roadmap graph, if neither of its two nodes is within the obstacles; see Fig.\ref{fig1} for an illustration.

\underline{Second}, pick the shortest path in the roadmap graph as the initial path, using Dijkstra’s algorithm.

\underline{Third}, apply the CFS algorithm to reshape the initial path until we reach the final path.

In this paper, each obstacle is approximated by square grids to reduce the calculation costs of convex feasible sets of the CFS algorithm. Especially, each obstacle is first dilated to fill the grids (dark blue in Fig.\ref{fig1}) that are overlapped with this obstacle and further dilated at least $\lceil d_{min}/\delta\rceil\delta$ (light blue region in Fig.\ref{fig1}) to meet the condition of minimum safety distance $d_{min}$, where $\lceil d_{min}/\delta\rceil\delta$ represents the smallest integer not less than $d_{min}/\delta$. Because the size of each grid is small enough, the size of each obstacle is not overestimated much.

In the rest of this paper, we refer to the dilated obstacles (the light blue region in Fig.\ref{fig1}) as the ones that we need to detour. It is worth noting that the distances from the initial waypoints (black dots in Fig.\ref{fig1}) to dilated obstacles are at least $\delta$, since every node of arcs in the roadmap graph is outside dilated obstacles. This setting guarantees the existence of nonempty convex feasible sets; see discussions below.

To prove that this new algorithm can guarantee to find a feasible path, we prove the following theorem based on Lemma \ref{lemma1} below.

\newtheorem{theorem}{Theorem}
\newtheorem{lemma}[theorem]{Lemma}
\begin{lemma}[Theorem 1 in \cite{ref20}]
	\label{lemma1}
	For every initial path whose convex feasible sets have nonempty interiors, the CFS algorithm will converge to a strong local optimum or weak local optimum of the optimization problem \eqref{equ1}-\eqref{equ3}.
\end{lemma}

\begin{theorem}
	If the whole configuration space is gridded with a sufficient small resolution level and obstacles are dilated as mentioned above, there must exist nonempty convex feasible sets for each waypoint of the initial path $\pmb{x}^{(0)}=[\pmb{x}_0^{(0)T},\pmb{x}_1^{(0)T},\dots,\pmb{x}_n^{(0)T}]^T$ found by Dijkstra’s algorithm.
\end{theorem}

\begin{figure}[htp]
	\centering
	\includegraphics[width=2.5in]{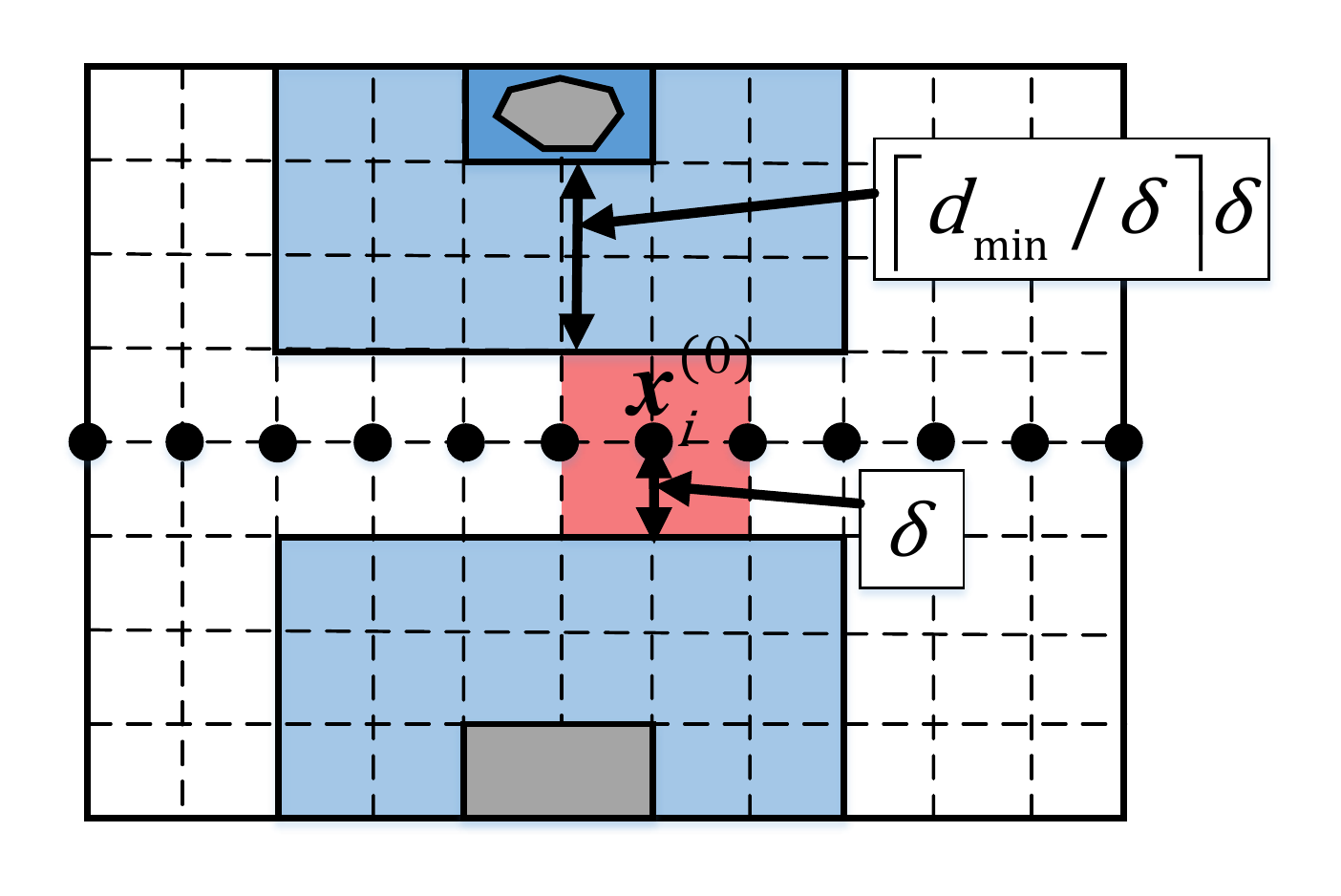}
	\caption{An illustration that the four grids connecting $\pmb{x}_i^{(0)}$ do not contain dilated obstacles (light blue regions).}
	\label{fig2}
\end{figure}

\begin{IEEEproof}
	We prove it by contradiction. Suppose the convex feasible sets $F(\pmb{x}^{(0)})=\emptyset$, there exists a waypoint $\pmb{x}_i^{(0)}$ with $F(\pmb{x}_i^{(0)})=\emptyset$ at least. Without loss of generality, we assume that $\pmb{x}_i^{(0)}$ does not lie in the boundaries of the configuration space. The situation where $\pmb{x}_i^{(0)}$ locates at the boundaries can be proved similarly.
	
	As the intersection point of two gridlines that are not boundaries, $\pmb{x}_i^{(0)}$ is the common vertex of four grids, as shown in Fig.\ref{fig2}. Due to $F(\pmb{x}_i^{(0)})=\emptyset$, there will exist dilated obstacles in an arbitrarily small neighborhood $\Big\{\pmb{x}\Big\lvert\big\lVert\pmb{x}-\pmb{x}_i^{(0)}\big\rVert\le\epsilon\Big\}$ of $\pmb{x}_i^{(0)}$.
	
	Since $\pmb{x}_i^{(0)}$ is a node of some arc in the roadmap graph, whose distance to the closest dilated obstacle is at least $\delta$, we can assert that the four neighboring grids connecting $\pmb{x}_i^{(0)}$ do not contain dilated obstacles at all. This contradicts the aforementioned result that at least one grid of the four grids contains dilated obstacles. Therefore, we have $F(\pmb{x}^{(0)})\neq\emptyset$.
\end{IEEEproof}

\subsection{Time Complexity Analysis and a Modified Algorithm}
\label{subsection3_2}
It is apparent that the time cost of this new algorithm consists of two parts. The first part comes from the calculation of the obstacle dilation and finding the shortest path. The second part comes from the iteration of the CFS algorithm.

Suppose we obtain $k\times l$ grids for the configuration space (that is, a roadmap graph with $k\times l$ nodes and $4k\times l$ arcs) and the found shortest path contains $(n+1)$ waypoints. Generally, for a rectangle configuration space, $(n+1)$ should be at the same order of $\sqrt{k\times l}$. The average time complexity for Dijkstra’s algorithm to find the shortest path in this roadmap graph is $O\big((k\times l)\textrm{log}(k\times l)+4k\times l\big)$. The dilation time cost is relatively small and is thus omitted here.

With $(n+1)$ waypoints, we have $2(n+1)$ decision variables for the optimization problem \eqref{equ1}-\eqref{equ3}. If we use the interior-point method \cite{ref25} (e.g., Mehrotra-type predictor-corrector algorithm \cite{ref26} used in MATLAB), the average time complexity of solving the optimization problem \eqref{equ1}-\eqref{equ3} should be $O\big(16(n+1)^4\lvert \textrm{log}(\epsilon)\rvert\big)$ \cite{ref26}, with the precision requirement $\epsilon$.

It is hard to predict how many iterations are needed for the CFS algorithm to converge. However, we can see that the time complexity of the whole algorithm is dominated by the time complexity of the CFS algorithm.

A further look shows that the CFS algorithm may works slowly if the number of waypoints in the initial path is too large. To solve this problem, we propose a modified algorithm that uses the divide-and-conquer method.

Beginning from the first waypoint, we divide the whole initial path into segments of every $m$ waypoints. With $(n+1)$ initial waypoints, we will have $\lceil (n+1)/m\rceil$ segments in the path and we will reshape each segment by using the CFS algorithm, respectively.

\begin{figure*}[htp]
	\centering
	\subfloat[]{\includegraphics[width=2.5in]{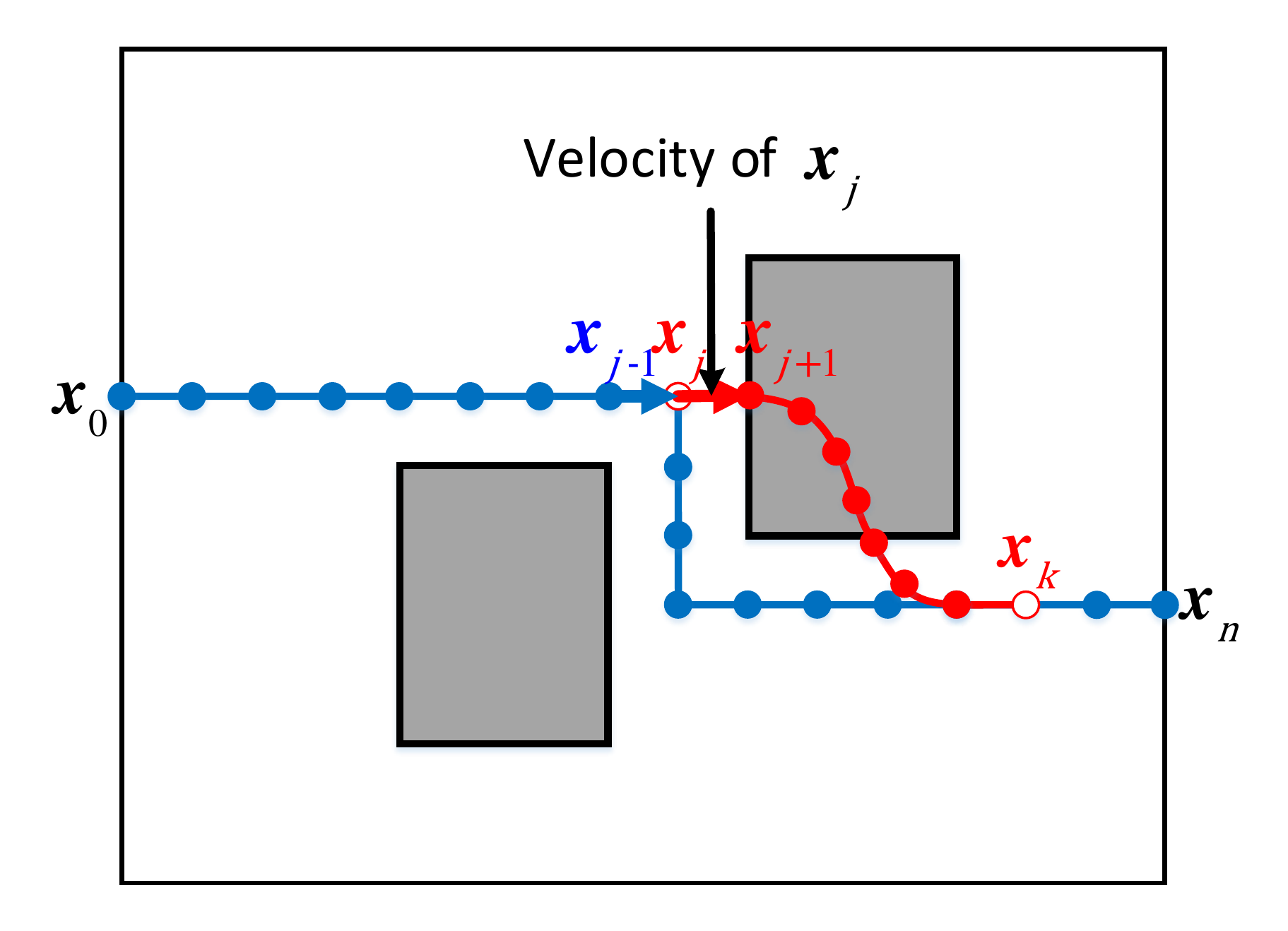}%
		\label{fig3_a}}
	\hfil
	\subfloat[]{\includegraphics[width=2.5in]{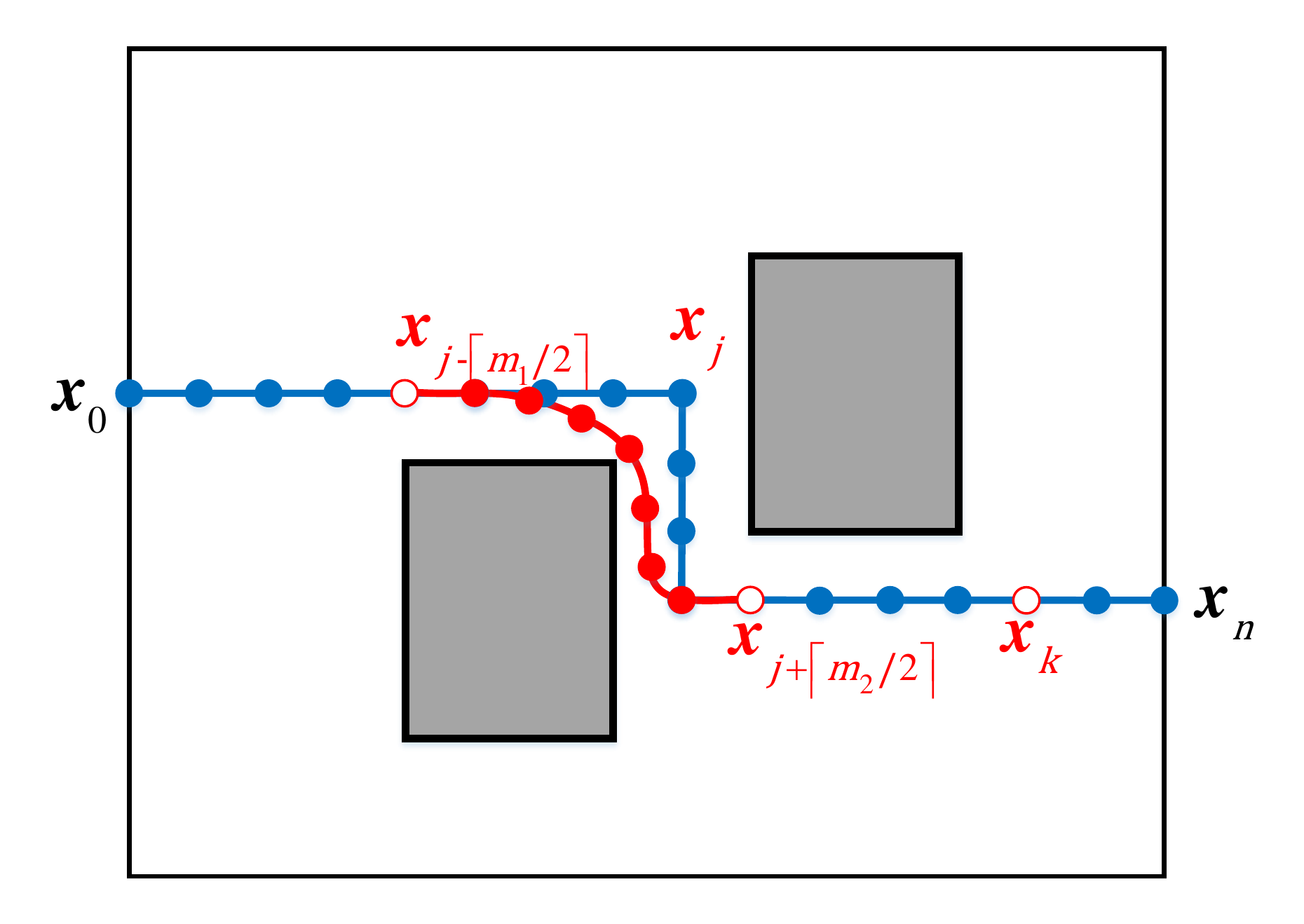}%
		\label{fig3_b}}
	\caption{(a) The boundary point $\pmb{x}_j$ is so close to obstacles that the reshaped segment (red dots) is inside the obstacle; (b) the boundary point $\pmb{x}_j$ is substituted by two new boundary points $\pmb{x}_{j-\lceil m_1/2\rceil}$ and $\pmb{x}_{j+\lceil m_2/2\rceil}$}
	\label{fig3}
\end{figure*}

This simple trick may fail, since there might exist some sharp turns around the boundary points of these segments. To avoid the abrupt changes of speed caused by these sharp turns, we append an extra constraint that the velocities of boundary points should be equal for two consecutive segments, as the velocities of their predecessors when using the CFS algorithm to reshape every segment. However, this additional constraint sometimes makes the CFS algorithm unable to obtain a feasible segment.

Fig.\ref{fig3_a} provides an intuitive illustration. In Fig.\ref{fig3_a}, $\pmb{x}_j$ is a boundary point, $\pmb{x}_{j-1}$ and $\pmb{x}_{j+1}$ are its predecessor and successor waypoint, respectively. Since the speed of $\pmb{x}_{j-1}$ is $[\delta,0]^T$ in Fig.\ref{fig3_a}, the speed of $\pmb{x}_j$ will be also set as $[\delta,0]^T$ according to the extra constraint. Moreover, because the distance from $\pmb{x}_j$ to the obstacle along the speed direction of $\pmb{x}_j$ is exact $\delta$, the continuity requirement of speed will drive $\pmb{x}_{j+1}$ to lie in the boundary of the obstacle, and the following waypoints will be inside the obstacle.

To solve this problem, we can select two waypoints $\pmb{x}_{j-\lceil m_1/2\rceil}$ and $\pmb{x}_{j+\lceil m_2/2\rceil}$ as new boundary points to replace $\pmb{x}_j$, where $m_1$ and $m_2$ are the number of waypoints in two segments that connect $\pmb{x}_j$, respectively, as shown in Fig.\ref{fig3_b}. If the boundary point still does not work, we can repeat such boundary point alternation until that the CFS algorithm obtains a feasible final segment. With the aid of the boundary point alternation, the CFS algorithm will have a larger probability (may not always 100\%) to obtain a feasible segment.

\begin{algorithm}
	\caption{Roadmap Path Reshaping-m (RPR-m) Algorithm}
	\label{algorithm2}
	\begin{algorithmic}[1]
		\STATE\pmb{Initinalize}\\
		\STATE\quad Set the number of grids: $k\times l$.\\
		\STATE\quad Set the number of waypoints in each segment: $m$.\\
		\STATE\quad Grid the whole configuration space.
		\STATE\quad Dilate obstacles at least $\big(\lceil d_{min}/\delta\rceil+1\big)\delta$.\\
		{\color{green} \textit{//divide the initial path into several segments}}
		\STATE Apply Dijkstra to find a collision-free initial path.\\
		\STATE Divide the initial path into $d=\lceil(n+1)/m\rceil$ segments and record all boundary points.\\
		{\color{green} \textit{//apply the CFS algorithm to modify every segment}}
		\STATE flag=0.\\
		\STATE\pmb{Loop}\\
		\STATE\quad Apply the CFS algorithm to modify the $i$th segment.\\
		{\color{green} \textit{//if CFS fails, then alter boundary points}}
		\STATE\quad\pmb{If} CFS fails to modify it\\
		\STATE\qquad\pmb{If} the number of waypoints of this segment over 3\\
		\STATE\qquad\quad Alter the $(i-1)$th boundary point.\\
		\STATE\qquad\quad $d=d+1,i=i-1$.\\
		\STATE\qquad\quad continue.\\
		\STATE\qquad\pmb{Else}
		\STATE\qquad\quad flag=1.\\
		\STATE\qquad\quad break.\\
		\STATE\qquad\pmb{End If}
		\STATE\quad\pmb{End If}\\
		\STATE\quad $i=i+1$.\\
		\STATE\pmb{Until} $i\ge d$\\
		{\color{green} \textit{//return the final path}}
		\STATE\pmb{If} flag=0\\
		\STATE\quad Connect all segments to generate the complete path.\\
		\STATE\quad\pmb{Return} the complete path.\\
		\STATE\pmb{Else}\\
		\STATE\quad\pmb{Return} the initial path.\\
		\STATE\pmb{End If} 
	\end{algorithmic}
\end{algorithm}

The modified algorithm can be summarized as Algorithm \ref{algorithm2} below. Testing results show that the modified algorithm usually works in practice. In the below, we use the abbreviation RPR-m to refer this algorithm with at most $m$ waypoints in a segment.

We generally neglect the time cost of the boundary point alternation due to the low probability of occurrence. Thus, the modified algorithm usually solves $(n+1)/m$ optimization subproblems \eqref{equ1}-\eqref{equ3} with $2m$ decision variables for each subproblem (assume that $(n+1)/m$ is an integer). The time complexity of solving each subproblem is $O\big(16m^4\lvert \textrm{log}(\epsilon)\rvert\big)$. Therefore, the total time complexity should be $O\big(16m^3(n+1)\lvert \textrm{log}(\epsilon)\rvert\big)$, which is much less than the complexity of directly solving the original optimization problem with $2(n+1)$ decision variables.
\section{Further Discussions On Curvature Constraints}
\label{section4}
Different with the CFS algorithm, our new algorithm can be further extended to explicitly consider the curvature constraints between two consecutive waypoints
\begin{equation}
\theta(\pmb{x}_i-\pmb{x}_{i-1},\pmb{x}_{i-1}-\pmb{x}_{i-2})\le\theta_{max},~i=2,3,\dots,n-2.
\label{equ5}
\end{equation}
where $\theta(\pmb{x}_i-\pmb{x}_{i-1},\pmb{x}_{i-1}-\pmb{x}_{i-2})$ represents the intersection angle between the vector $(\pmb{x}_i-\pmb{x}_{i-1})$ and $(\pmb{x}_{i-1}-\pmb{x}_{i-2})$, and $\theta_{max}$ is the maximum permittable intersection angle.

\subsection{Considering Curvature Constraints at Roadmap Step}
\label{subsection4_1}
The first extension is to handle the curvature constraints in the roadmap step by applying the Beamlet-based Dijkstra’s algorithm \cite{ref27}, named the Beamlet algorithm, instead of Dijkstra’s algorithm to find the initial path.

\begin{figure}[htp]
	\centering
	\includegraphics[width=2.5in]{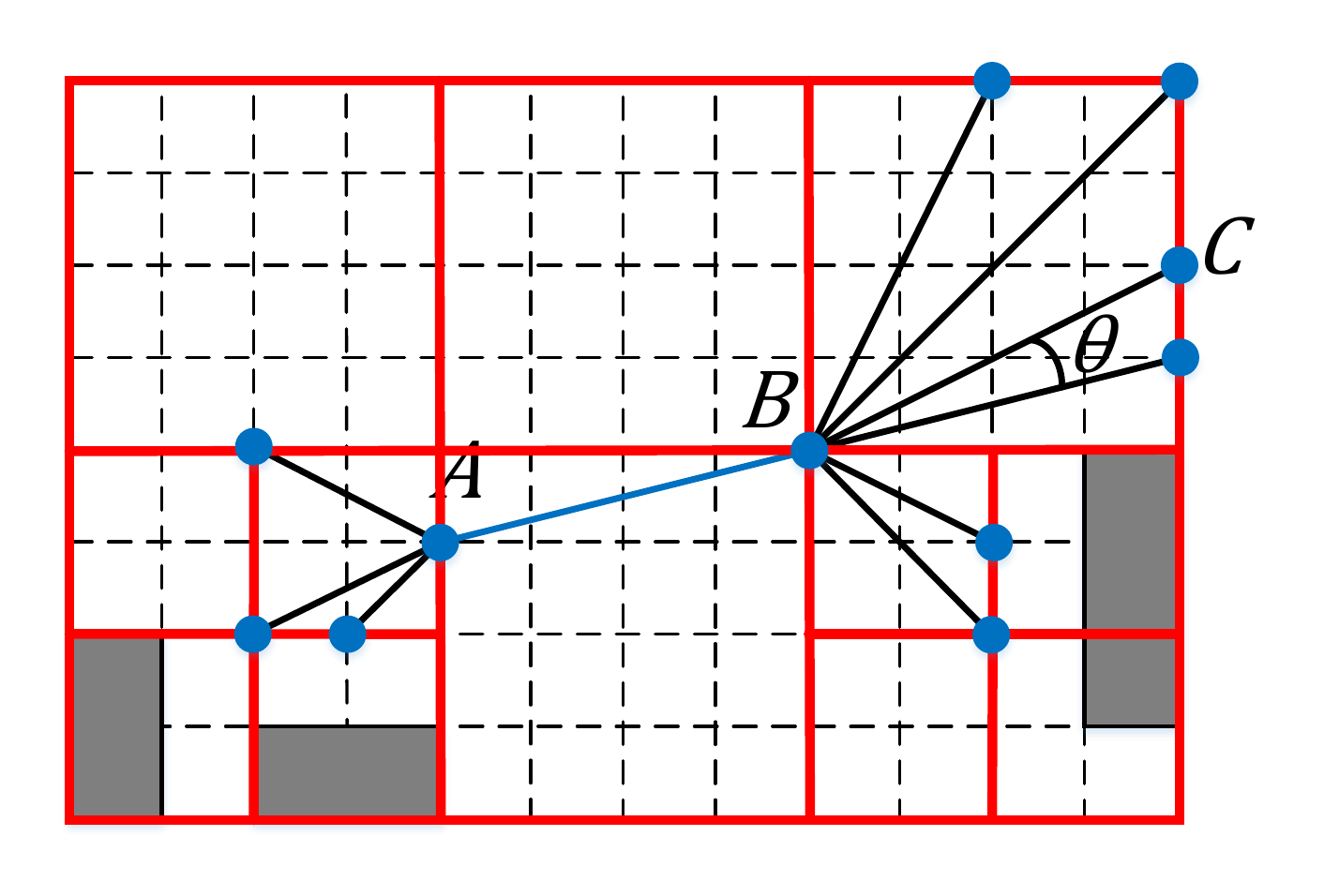}
	\caption{An illustration of beamlets and dyadic squares (red squares), where grey polygons represent obstacles.}
	\label{fig4}
\end{figure}

Beamlets refer to the line segments (e.g., the line segment AB in Fig.\ref{fig4}) connecting certain points located at the boundaries of dyadic squares. A dyadic square \cite{ref28} (d-square, e.g., the red squares in Fig.\ref{fig4}) is a set of points in a square region with pre-specified size. A d-square may be further partitioned into four sub-d-squares of the same size to guarantee that beamlets constructed on these d-squares do not pass through obstacles.

In this paper, we select the gridlines of the roadmap constructed in Algorithm \ref{algorithm2} as the boundaries of dyadic squares. Furthermore, the nodes of the roadmap that locate at the boundaries of dyadic squares are taken as endpoints of beamlets.

We can further construct a beamlet graph by linking a series of beamlets, with respect to the given curvature constraints. The intersection angle between the beamlet AB and the beamlet BC is defined as the intersection angle between the vector $\overrightarrow{AB}$ and $\overrightarrow{BC}$ ($\theta(\overrightarrow{AB},\overrightarrow{BC})$), i.e. $\theta$ in Fig.\ref{fig4}. In a beamlet graph \cite{ref28}, an arc $e(AB,BC)$ exists from the beamlet AB to the beamlet BC, if and only if they are connected and $\theta(\overrightarrow{AB},\overrightarrow{BC})\le\theta_{max}$. Notably, the length of $e(AB,BC)$ is set as the length of the beamlet BC.

The Beamlet algorithm first constructs a beamlet graph based on the original roadmap in Algorithm \ref{algorithm2} and then uses Dijkstra’s algorithm on the beamlet graph to find an initial path. More details can be found in \cite{ref27} -\cite{ref28}.

Notably, if some beamlet is too long, the distance between corresponding waypoints (the endpoints of the beamlet) may be also too large. To balance the distance between consecutive waypoints, we suggest inserting extra waypoints along the initial path found by the Beamlet algorithm.

\subsection{Considering Curvature Constraints at Reshaping Step}
\label{subsection4_2}
The second extension is to use an iterative optimization algorithm to make a given path rigorously meet the curvature constraints.

\begin{figure}[htp]
	\centering
	\includegraphics[width=2.5in]{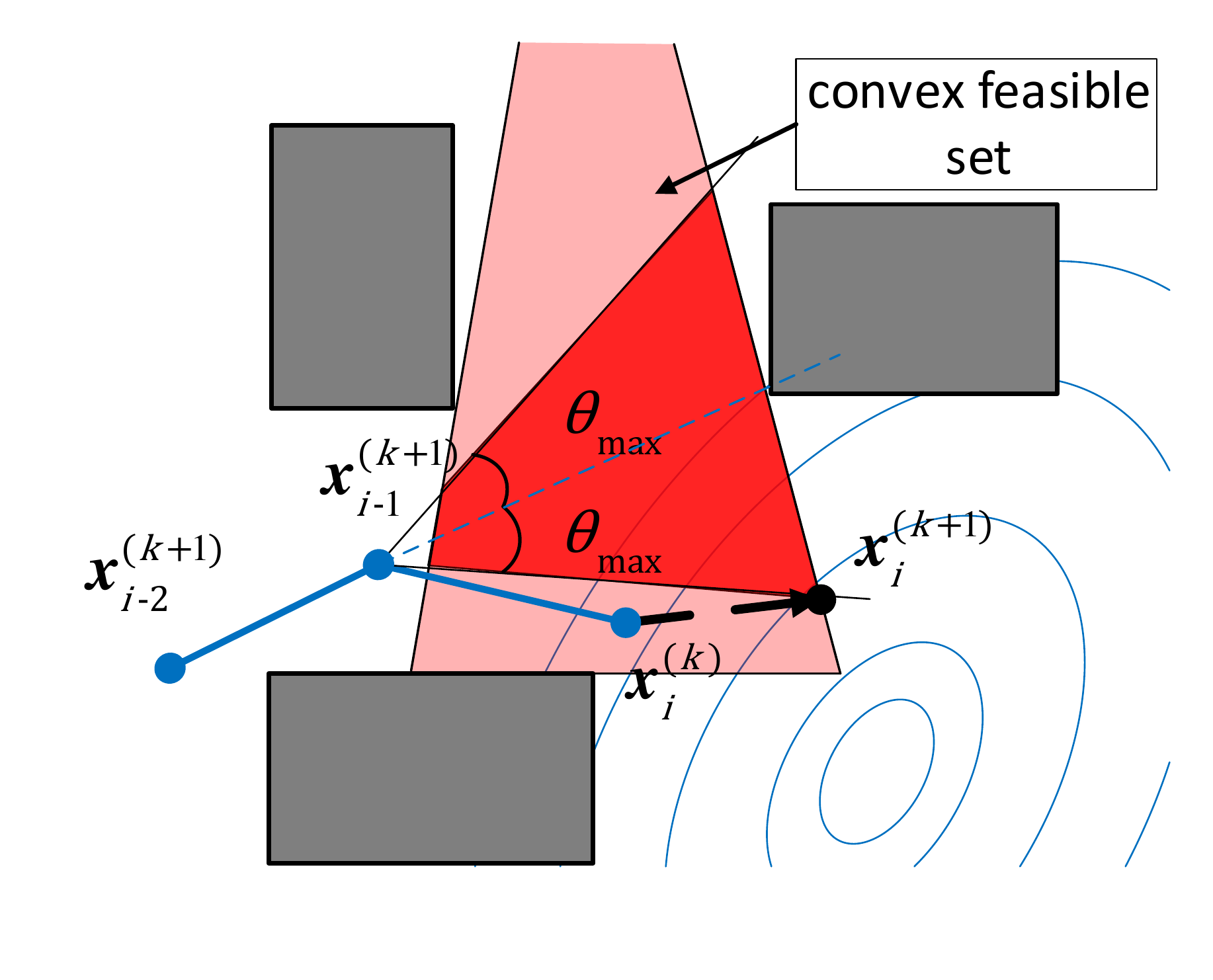}
	\caption{An illustration of an iteration, where the light red region is the convex feasible set and the dark red region is the feasible region of the optimization problem \eqref{equ6}-\eqref{equ9}. The grey polygons represent obstacles and the blue contour represents the cost function \eqref{equ6}.}
	\label{fig5}
\end{figure}

Specifically, in this paper, we only consider the situation where $\theta_{max}<\frac{\pi}{2}$. Denoting $\pmb{x}_i=[a_i,b_i]^T$, we iteratively solve the following optimization problem \eqref{equ6}-\eqref{equ9} to sequentially check and modify every waypoint $\pmb{x}_i$ while fixing other waypoints $\pmb{x}_j, j\neq i$, until the whole path meets these constraints or the number of modification is over a pre-specified value; see Fig.\ref{fig5} for an illustration.

Since only $\pmb{x}_i$ is modified and other waypoints are unchanged, the constraint \eqref{equ9} becomes convex. Consequently, the obtained optimization problem \eqref{equ6}-\eqref{equ9} is convex.

\begin{align}
\label{equ6}\max_{\pmb{x}_i}\quad &J(\pmb{x})=\pmb{x}(\pmb{V}^T\pmb{V}+\lambda\pmb{A}^T\pmb{A})\pmb{x}\\
\label{equ7}s.t.\quad & \pmb{x}_0=\pmb{x}_{start},~\pmb{x}_n=\pmb{x}_{end}\\
\label{equ8}&\pmb{x}\in F(\pmb{x})\\
\label{equ9}\begin{split}&c_1\big[(a_{i-2}-a_{i-1})a_i+(b_{i-2}-b_{i-1})b_i+a_{i-1}^2+\\
&b_{i-1}^2-a_{i-1}a_{i-2}-b_{i-1}b_{i-2}\big]+\big\lvert(a_{i-1}-a_{i-2})b_i\\
&-(b_{i-1}-b_{i-2})a_i+b_{i-1}a_{i-2}-b_{i-2}a_{i-1}\big\rvert\le 0\end{split}
\end{align}
where $F(\pmb{x})$ is the convex feasible set and $c_1=\textrm{tan}\theta_{max}$. The constraint \eqref{equ9} is equivalent to $\textrm{tan}~\theta_i\le c_1,~i=2,3,\dots,n-2$.

Our extended algorithm can be summarized as Algorithm \ref{algorithm3} below. In the below, we use the abbreviation ERPR-m to refer this extended algorithm with at most $m$ waypoints in a segment.

\begin{algorithm}
	\caption{Extended Roadmap Path Reshaping-m (RPR-m) Algorithm}
	\label{algorithm3}
	\begin{algorithmic}[1]
		\STATE\pmb{Initinalize}\\
		\STATE\quad Set the number of grids: $k\times l$.\\
		\STATE\quad Set the number of waypoints in each segment: $m$.\\
		\STATE\quad Grid the whole configuration space.
		\STATE\quad Dilate obstacles at least $\big(\lceil d_{min}/\delta\rceil+1\big)\delta$.\\
		\STATE\quad Set the maximum number of iterations: maxiter.\\
		{\color{green} \textit{//construct the initial path that meets curvature constraints}}
		\STATE Apply the Beamlet algorithm to find an initial path.\\
		\STATE Insert extra waypoints if necessary.\\
		\STATE Divide the initial path into $d=\lceil(n+1)/m\rceil$ segments and record all boundary points.\\
		\STATE Apply the CFS algorithm to reshape every segment.\\
		
		{\color{green} \textit{//iteratively adjust the curvature of the path}}
		\STATE $i=0$.\\
		\STATE\pmb{Loop}\\
		\STATE\quad Solve the optimization problem \eqref{equ6}-\eqref{equ9} to modify every waypoint sequentially.\\
		\STATE\quad $i=i+1$.\\
		\STATE\pmb{Until} the whole path meets our goal \pmb{or} $i>$maxiter.\\
		{\color{green} \textit{//return the final path}}
		\STATE\pmb{If} successfully modify the whole path\\
		\STATE\quad\pmb{Return} the complete path.\\
		\STATE\pmb{Else}\\
		\STATE\quad\pmb{Return} the initial path.\\
		\STATE\pmb{End If}.
	\end{algorithmic}
\end{algorithm}

\subsection{Time Complexity Analysis}
\label{subsection4_3}
It is apparent that Algorithm \ref{algorithm3} has two additional time costs than Algorithm \ref{algorithm2}.

First, Algorithm \ref{algorithm3} applies Dijkstra’s algorithm on the beamlet graph to find the initial path. According to \cite{ref27}, with $k\times l$ nodes in the roadmap, there will be $O\big(\frac{1}{2}(k\times l)\textrm{log}_2(k\times l)\big)$ nodes and $O\Big(\frac{1}{4}(k\times l)^2\big(\textrm{log}_2(k\times l)\big)^3\Big)$ arcs in the beamlet graph that is based on the roadmap. Therefore, the time complexity for Dijkstra’s algorithm to find initial paths on the beamlet graph is $O\Big(\frac{1}{4}(k\times l)^2\big(\textrm{log}_2(k\times l)\big)^3\Big)$, which can be rewritten as $O\Big(2(n+1)^4\big(\textrm{log}_2(n+1)\big)^3\Big)$, since $k\times l$ is at the same order of $(n+1)^2$, the square of the number of waypoints, for rectangle configuration spaces. Here we ignore the complexity of constructing the beamlet graph, since it is much less.

Second, Algorithm \ref{algorithm3} needs to check and modify each waypoint, corresponding to solving the additional optimization problem \eqref{equ6}-\eqref{equ9}. We still assume that Mehrotra-type predictor-corrector algorithm is used to solve it, whose average time complexity is $O\big(n^4\lvert \textrm{log}(\epsilon)\rvert\big)$ if there are $n$ decision variables and the precision requirement is $\epsilon$. Since the optimization problem \eqref{equ6}-\eqref{equ9} only has 2 decision variables, the complexity of modifying a waypoint is $O\big(16\lvert \textrm{log}(\epsilon)\rvert\big)$. If the path has totally $(n+1)$ waypoints, then the complexity of modifying the whole path is  $O\big(16(n+1)\lvert \textrm{log}(\epsilon)\rvert\big)$.

\section{Numerical Tesing Results}
\label{section5}
\subsection{Comparison between the CFS and the Algorithm \ref{algorithm2}}
\label{subsection5_1}
In this subsection, the performance of Algorithm \ref{algorithm2} (RPR-m algorithm) is compared with the original CFS algorithm on random configuration spaces.

We restrict the whole configuration space in a rectangular region whose length and width are 9 and 6, respectively, and set up a coordinate system as shown in Fig.6. In addition, we set $d_{min}=0.1$ and discretize the configuration space into $60\times 90$ grids with the resolution level $\delta=0.1$.

We randomly place non-overlapping rectangular obstacles in the configuration space \cite{ref29}, whose position satisfies the 2D uniform distribution in the configuration space and aspect ratio satisfies the 1D uniform distribution in the interval [1/2.5,2.5]. The area of the $i$th rectangle is set as $A_0/i^{1.1}$ \cite{ref29}, where $A_0$ is pre-specified to guarantee that these rectangles can fill the whole configuration space if their number is infinite.

For the motion planning problem (1)-(3), we set $\lambda=1$, $\pmb{x}_{start}=[0,0]^T$, $\pmb{x}_{end}=[9,0]^T$. The other parameters of the RPR-m algorithm and the original CFS algorithm are set as follows:
\begin{itemize}
	\item The RPR-m algorithm. We set $m=60$, considering the capability of our computer and the fact that a small $m$ might make the CFS algorithm inefficient because there are too many segments to reshape. We refer to it as RPR-60. Specially, we also use the abbreviation RPR-ALL to refer our algorithm that uses the CFS algorithm directly reshape the whole path without any segments.
	\item The original CFS algorithm. We set its initial path as the line segment bounded between the starting point and the ending point, and set $(n+1)$, the number of waypoints, to be equal as the number of waypoints of the initial path in the RPR algorithm.
\end{itemize}

We compare the performance of the RPR algorithm (including both RPR-ALL and RRP-60) and the original CFS algorithm in the following two criteria.
\begin{enumerate}
	\item The probability to find a feasible path.
	\item The computation time of the algorithm.
\end{enumerate}

To avoid bias caused by occasionality, we randomly generated 1000 configuration spaces, which can be divided into 5 groups. There are 200 configuration spaces in each group. The configuration spaces in each group contain 5, 10, 15, 20, and 30 obstacles, respectively.

All numerical experiments were performed on a computer with an Intel(R) Core (TM) i7-7700U CPU and 8GB RAM. The RPR algorithm and the original CFS algorithm were implemented in MATLAB2016b. We implement the CFS algorithm according to Liu \cite{ref30}. The main differences between our code and Liu’s code are that we add additional constraints to strict paths in the configuration space and the relative optimization problems were solved by CVX \cite{ref31}.

Table.\ref{table1} shows the detailed results of these algorithms. The third column of Table.\ref{table1} shows the number and the percentage of problems that have been solved successfully by each algorithm, respectively. Clearly, both RPR-ALL and RPR-60 can find a feasible path with 100\% probability; while the original CFS algorithm fails to do so especially when the number of obstacles is large.

The fourth column of Table.\ref{table1} shows the average computation time of each algorithm. In particular, the numbers in parentheses represent the average computation time of finding initial paths. Since the initial path of the original CFS algorithm is the line segment bounded between the starting point and the ending point, we assume that its computation time is zero. For every group, the computation time for the RPR-ALL algorithm and the RPR-60 algorithm to find initial paths is nearly zero. Moreover, the RPR-ALL algorithm converges slightly faster than the original CFS algorithm but much slower than the RPR-60 algorithm. This is mainly caused by the proper partition of the whole path which may lead to a noticeable reduction of time costs.

The fifth column of Table.\ref{table1} shows the average number of waypoints in each group. It is obvious that the number of waypoints tends to increase as the number of obstacles increases, which is consistent with the intuition that more complicated environments usually lead to longer trajectories.

\begin{figure*}[!t]
	\centering
	\subfloat[]{\includegraphics[width=2.5in]{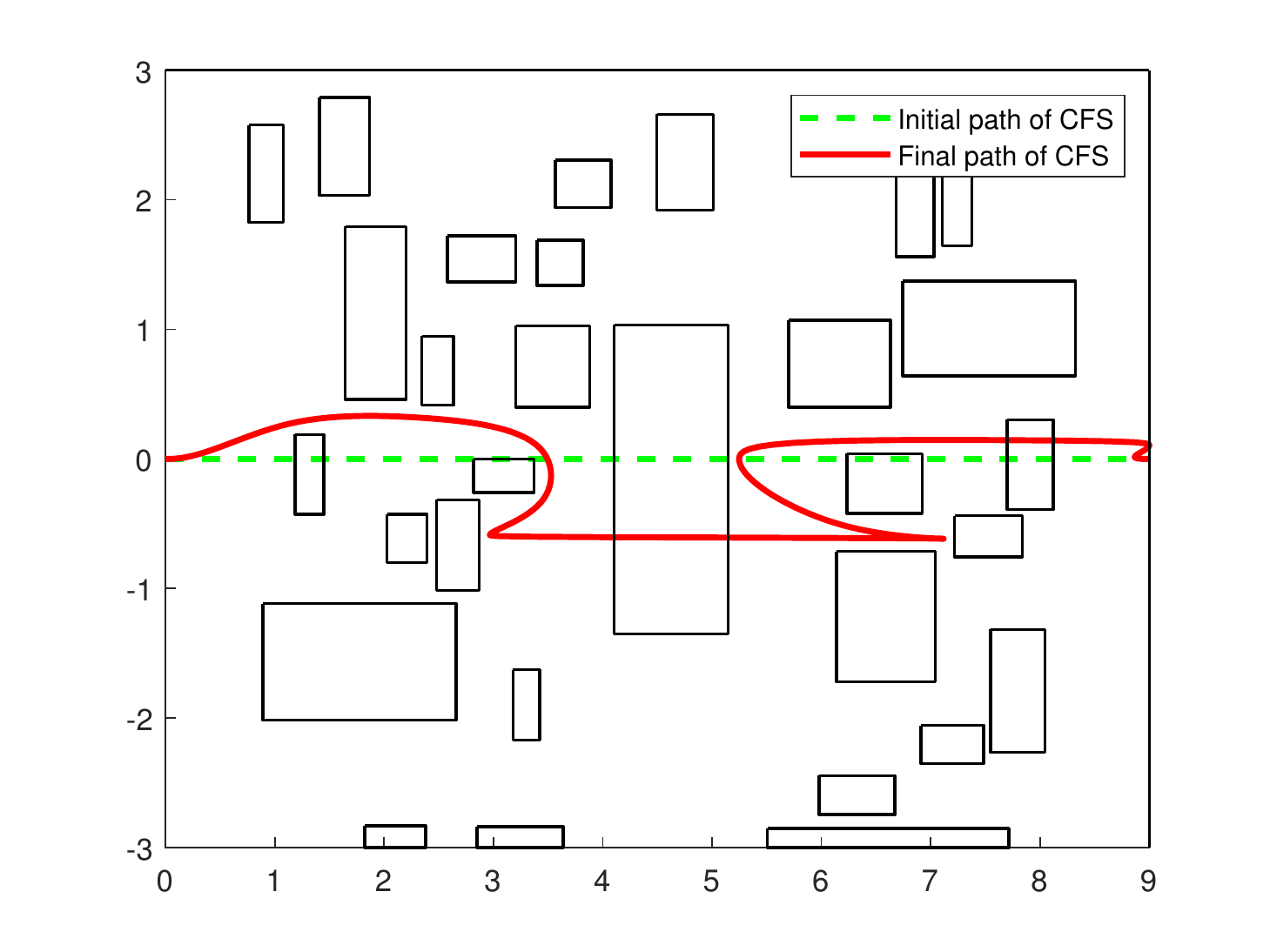}%
		\label{fig6_a}}
	\hfil
	\subfloat[]{\includegraphics[width=2.5in]{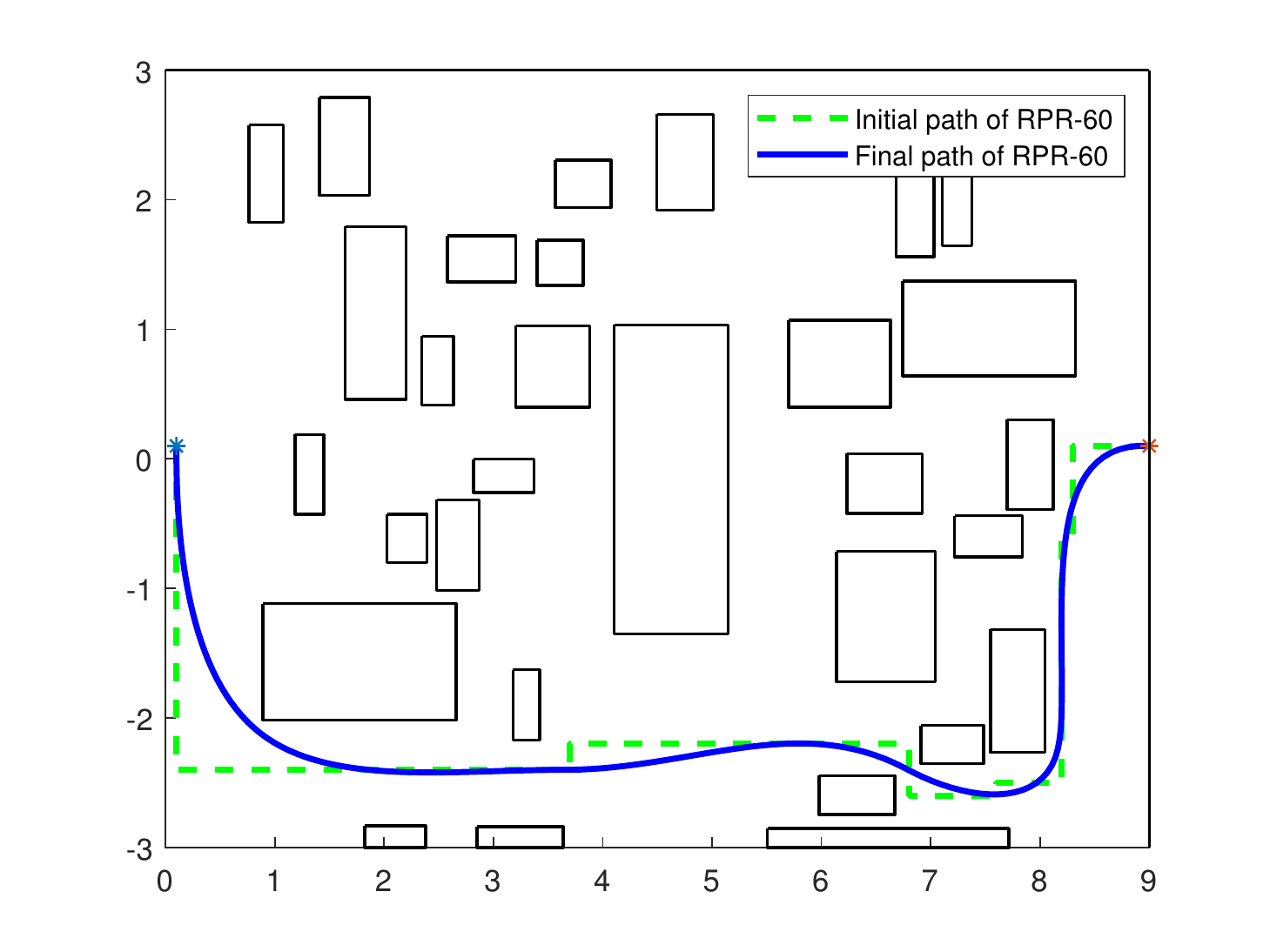}%
		\label{fig6_b}}
	\caption{(a) The randomly picked-up initial path (green dash curve) and the final path (red solid curve) computed by the CFS algorithm; (b) the initial path (green dash curve) and the final path (blue solid curve) computed by the RPR-60 algorithm.}
	\label{fig6}
\end{figure*}

To provide an intuitive illustration, Fig.\ref{fig6_a} and Fig.\ref{fig6_b} show an example of the final paths computed by the RPR-60 algorithm and the original CFS algorithm, respectively. We can see that the RPR-60 algorithm successfully found a feasible and smooth final path, while the original CFS algorithm failed.

In summary, compared to the original CFS algorithm, the RPR algorithm requires less computation time and can almost surely find a feasible and smooth path.

\begin{table}[htbp]
	\caption{Performance of RPR-ALL, RPR-60, and CFS in 1000 Random Configuration Spaces}
	\label{table1}
	\setlength{\tabcolsep}{3pt}
	\begin{tabular}{|p{28pt}|p{35pt}|p{65pt}|p{57pt}|p{30pt}|}
		\hline
		Group & Method&How many configuration spaces that can be solved&Average total time (average time of finding initial paths)& The average number of waypoints.\\
		\hline
		\multirow{3}{*}{Group1}&CFS&150/200(75\%)&2.17s(0s)&108\\
		\cline{2-5}
		&RPR-ALL&\pmb{200/200(100\%)}&1.75s(0.001s)&108\\
		\cline{2-5}
		&RPR-60&\pmb{200/200(100\%)}&1.46s(0.001s)&108\\
		\hline
		\multirow{3}{*}{Group2}&CFS&121/200(60.5\%)&4.75s(0s)&115\\
		\cline{2-5}
		&RPR-ALL&\pmb{200/200(100\%)}&4.27s(0.001s)&115\\
		\cline{2-5}
		&RPR-60&\pmb{200/200(100\%)}&1.98s(0.001s)&115\\
		\hline
		\multirow{3}{*}{Group3}&CFS&89/200(44.5\%)&13.71s(0s)&125\\
		\cline{2-5}
		&RPR-ALL&\pmb{200/200(100\%)}&10.37s(0.001s)&125\\
		\cline{2-5}
		&RPR-60&\pmb{200/200(100\%)}&2.35s(0.001s)&125\\\hline
		\multirow{3}{*}{Group4}&CFS&79/200(34.5\%)&24.64s(0s)&129\\
		\cline{2-5}
		&RPR-ALL&\pmb{200/200(100\%)}&18.29s(0.001s)&129\\
		\cline{2-5}
		&RPR-60&\pmb{200/200(100\%)}&3.11s(0.001s)&129\\\hline
		\multirow{3}{*}{Group5}&CFS&33/200(16.5\%)&64.87s(0s)&141\\
		\cline{2-5}
		&RPR-ALL&\pmb{200/200(100\%)}&56.8s(0.001s)&141\\
		\cline{2-5}
		&RPR-60&\pmb{200/200(100\%)}&5.93s(0.001s)&141\\	
		\hline			
	\end{tabular}
\end{table}

\subsection{Comparison between the Algorithm \ref{algorithm2} and the Algorithm \ref{algorithm3}}
\label{subsection5_2}
In this subsection, we compare Algorithm \ref{algorithm3} (ERPR-m algorithm) with Algorithm \ref{algorithm2} (RPR-m algorithm) in the perspective of the probability to find feasible paths that meet the curvature constraints \eqref{equ5}.

\begin{figure*}[!t]
	\centering
	\subfloat[]{\includegraphics[width=2.5in]{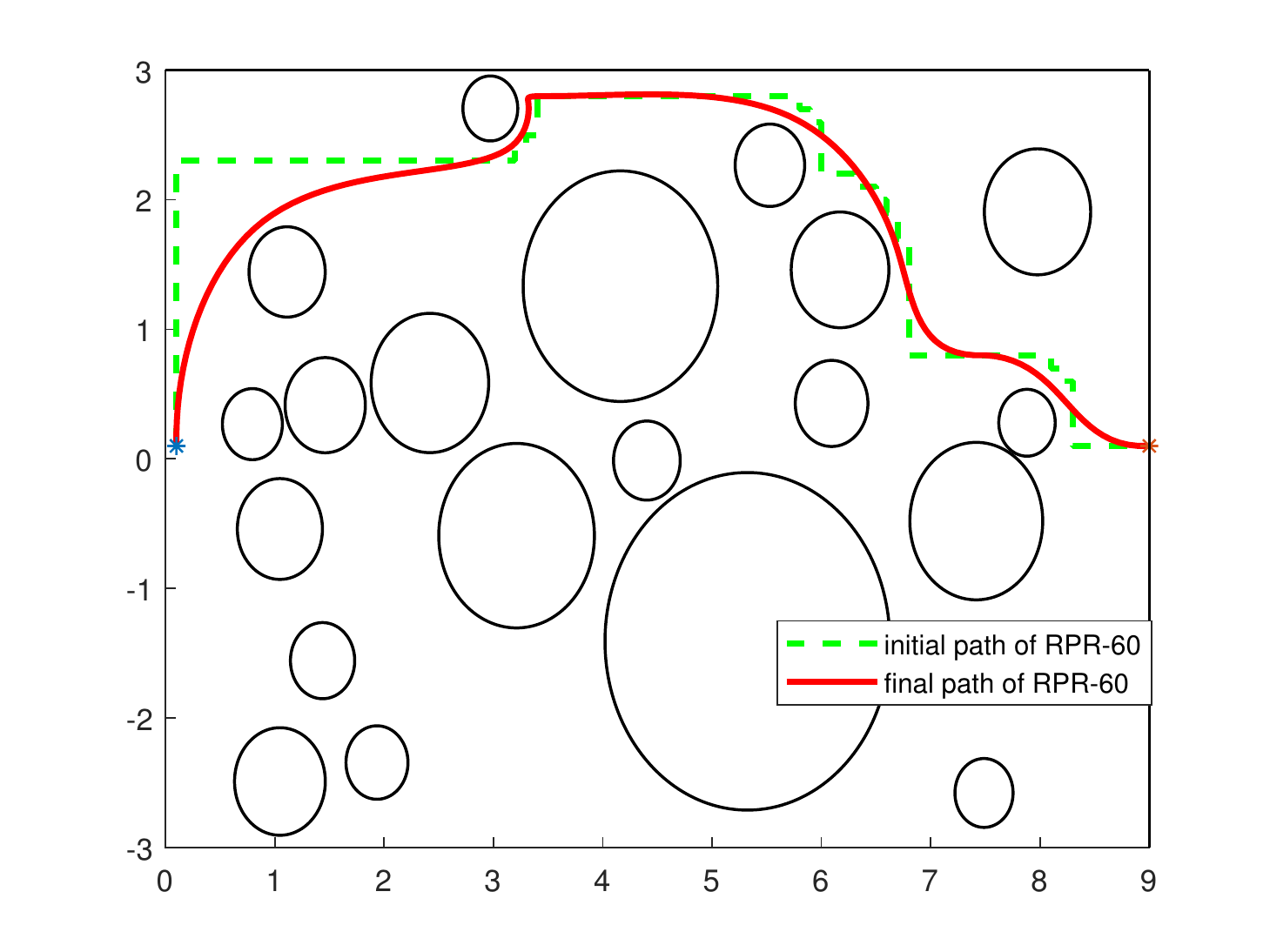}%
		\label{fig7_a}}
	\hfil
	\subfloat[]{\includegraphics[width=2.5in]{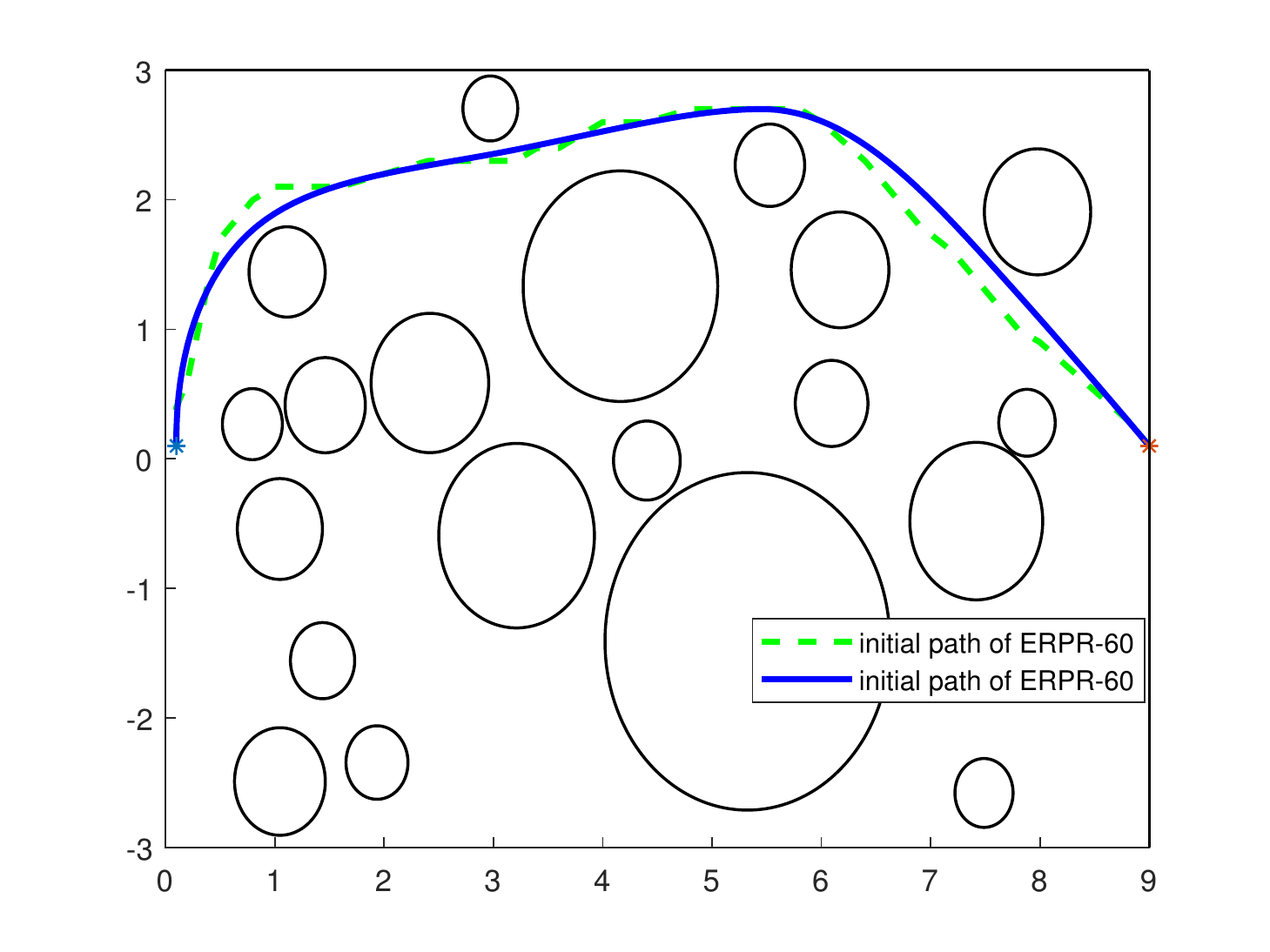}%
		\label{fig7_b}}
	\caption{(a) The initial path (green dash curve) and the final path (red solid curve) computed by the RPR-60 algorithm; (b) the initial path (green dash curve) and the final path (blue solid curve) computed by the ERPR-60 algorithm.}
	\label{fig7}
\end{figure*}

We still adopt the rectangular configuration spaces with the same size as the last subsection and set up the same coordinate system as shown in Fig.\ref{fig7}. The configuration space is also discretized into $60\times 90$ grids with the resolution level $\delta=0.1$ and $d_{min}=0.1$.

Here we randomly generate non-overlapping circular obstacles, whose position satisfies the 2D uniform distribution in the configuration space. The area of the $i$th circle is also set as $A_0/i^{1.1}$.

For the RPR-m algorithm, we set $m=60$; for the ERPR-m algorithm, we set $\theta_{max}=30^\circ$ and $m=60$. And we set “maxiter” of the ERPR-m algorithm as the number of waypoints, thus we can modify every waypoint in the path. The other parameters are the same as the last subsection.

We randomly generated 800 configuration spaces that are divided into 4 groups. There are 200 configuration spaces in each group, which contain 5, 10, 15, and 20 obstacles, respectively.

Table.\ref{table2} shows the detailed results of the two algorithms. The third column of Table.\ref{table2} shows the number and the percentage of problems that have been solved successfully by two algorithms, respectively. We can see that the ERPR-60 algorithm always has a larger probability to find feasible paths that meet curvature constraints, if compared to the RPR-60 algorithm. Of course, the probability for both the RPR-60 algorithm and the ERPR-60 algorithm to find feasible paths decreases as the number of obstacles increases.

The fourth column of Table.\ref{table2} shows the average computation time of each algorithm, where the numbers in parentheses represent the average computation time of finding initial paths. We can see that the average time for the ERPR-60 algorithm to reshape initial paths is similar to the corresponding time needed by the RPR-60 algorithm; while ERPR-60 algorithm requires more time to find initial paths. This is consistent with our previous time complexity analysis, since the complexity of the Beamlet algorithm is $O\Big(2(n+1)^4\big(\textrm{log}_2(n+1)\big)^3\Big)$, while the complexity of Dijkstra’s algorithm used in RPR-60 algorithm is $O\big((k\times l)\textrm{log}(k\times l)+4k\times l\big)$ (i.e. $O\big(2(n+1)^2\textrm{log}(n+1)+4(n+1)^2\big)$). The increase in computation time is acceptable, especially for cluttered configuration spaces, such as Group4, if we would like to increase the probability to find feasible paths.

Besides, we can see that the computation time for the Beamlet algorithm to find initial paths reduces with the number of obstacles increasing. As the number of obstacles increases, the total areas of d-squares that do not contain obstacles decreases, which contributes to a reduction in the number of the nodes and the arcs in the beamlet graph. Therefore, the time of finding initial paths on the beamlet graph reduces.

The fifth column of Table.\ref{table2} shows the average number of waypoints in each group. Obviously, the number of waypoints increases as the number of obstacles increases.

\begin{table}[htp]
	\caption{Performance of RPR-60, and ERPR-60 in 800 Random Configuration Spaces}
	\label{table2}
	\setlength{\tabcolsep}{4pt}
	\begin{tabular}{|p{28pt}|p{35pt}|p{65pt}|p{57pt}|p{30pt}|}
		\hline
		Group & Method&How many configuration spaces that can be solved&Average total time (average time of finding initial paths)& The average number of waypoints.\\
		\hline
		\multirow{2}{*}{Group1}&RPR-60&193/200(96.5\%)&1.34s(0.001s)&115\\
		\cline{2-5}
		&ERPR-60&\pmb{199/200(99.5\%)}&28.17s(27.09s)&88\\
		\hline
		\multirow{2}{*}{Group2}&RPR-60&177/200(88.5\%)&1.79s(0.001s)&129\\
		\cline{2-5}
		&ERPR-60&\pmb{191/200(95.5\%)}&13.99s(12.59s)&93\\
		\hline
		\multirow{2}{*}{Group3}&RPR-60&165/200(82.5\%)&2.06s(0.001s)&137\\
		\cline{2-5}
		&ERPR-60&\pmb{181/200(90.5\%)}&8.49s(6.50ss)&98\\
		\hline
		\multirow{2}{*}{Group4}&RPR-60&139/200(69.5\%)&2.70s(0.001s)&142\\
		\cline{2-5}
		&ERPR-60&\pmb{154/200(77\%)}&6.77s(3.70s)&102\\
		\hline			
	\end{tabular}
\end{table}

To provide an intuitive illustration, Fig.\ref{fig7_a} and Fig.\ref{fig7_b} show an example of the final paths found by the RPR-60 algorithm and the ERPR-60 algorithm, respectively. We can see that the ERPR-60 algorithm successfully found a feasible and smooth path that met the curvature constraints, while the final path found by the RPR-60 algorithm in Fig.\ref{fig7_a} is infeasible because there exists a sharp turn with a large angle, as shown in Fig.\ref{fig7_a}.

\section{Conclusion}
\label{section6}
Compared to the recently proposed CFS algorithm, our algorithm has a larger probability to find feasible paths that meet specific requirements, such as the curvature constraints, and achieves a great reduction in computation time. The key elements that contribute to the success of our algorithm stem from combining roadmap algorithms and the optimization-based path reshaping algorithms. Roadmap algorithms serve to find paths for discrete problems via rough but fast exploration, while the optimization-based path reshaping algorithms are designed to compute paths for continuous problems via refined and local adjustment. Therefore, combining the two kinds of algorithms can effectively balance the breadth and the depth of the search.

With the curvature constraint considered, the computation time will increase. This is reasonable and acceptable, as the old saying goes, “A beard well lathered is half shaved”.

In this paper, we simplify the shape of the robot into a single point, which is reasonable if the robot is much smaller than obstacles. However, in many applications, such as autonomous driving, the size of the robot is close to, even larger than the size of obstacles. In these situations, the shape of the robot should be considered, which will be handled in our future research.

\appendix
The main notations presented in this paper are summarized in the following Table.\ref{tableA1}.

\begin{table}[htp]
	\renewcommand{\arraystretch}{1.3}
	\caption{The Nomenclature List}
	\label{tableA1}
	\begin{tabular}{|p{70pt}|p{145pt}|}
		\hline
		Notation&Description\\
		\hline
		$\pmb{x}$&The whole path.\\
		\hline
		$\pmb{x}^{(k)}$&The whole path in iteration $k$.\\
		\hline
		$\pmb{x}_i$&The position of the robot at $i$th time in the configuration space.\\
		\hline
		$\pmb{x}_i^{(k)}$&The position of the robot at $i$th time in iteration $k$.\\
		\hline
		$\pmb{x}_{start},\pmb{x}_{end}$&The starting waypoint and the ending waypoint.\\
		\hline
		$[a_i,b_i]^T$&The coordinates of $\pmb{x}_i$.\\
		\hline
		$F(\pmb{x})$&The convex feasible sets.\\
		\hline
		$n+1$&The number of waypoints of the whole path.\\
		\hline
		$m$&The number of waypoints that each segment contains at most.\\
		\hline
		$O_i$&The $i$th obstacle.\\
		\hline
		$d(\pmb{x}_i,O_j)$&The distances between the robot and the $j$th obstacle in the $i$th time.\\
		\hline
		$d_{min}$&The minimum clearance between the robot and obstacles.\\
		\hline
		$k\times l$&The size of the roadmap.\\
		\hline
		$\delta$&The resolution level of the roadmap.\\
		\hline
		$\lambda$&Nonnegative coefficients in the objective function \eqref{equ1}.\\
		\hline
		$\theta(\pmb{x}_{i-1}-\pmb{x}_{i-2},\pmb{x}_{i}-\pmb{x}_{i-1})$&The intersection angle between the vector $(\pmb{x}_{i-1}-\pmb{x}_{i-2})$ and $(\pmb{x}_{i}-\pmb{x}_{i-1})$.\\
		\hline
		$\theta(\overrightarrow{AB},\overrightarrow{BC})$&The intersection angle between the vector $\overrightarrow{AB}$ and $\overrightarrow{BC}$.\\
		\hline
		$\theta_{max}$&The maximum permittable intersection angle.\\
		\hline
		$e(AB,BC)$&The arc of the beamlet graph, which connects the beamlet $AB$ and $BC$.\\
		\hline			
	\end{tabular}
\end{table}


\begin{thebibliography}{1}

\bibitem{ref1}
R. Geraerts, and M. H. Overmars, ``Creating high-quality paths for motion planning,'' \emph{The International Journal of Robotics Research}, vol. 26, no. 8, pp. 845--863, 1979.

\bibitem{ref2} J. -C. Latombe, ``Robot motion planning,'' \emph{Springer}, New York, vol. 124, 2012.

\bibitem{ref3} Z. Zhu, E. Schmerling, and M. Pavone, ``A convex optimization approach to smooth trajectories for motion planning with car-like robots,'' \emph{Decision and Control (CDC), 2015 IEEE 54th Annual Conference on}, pp. 835--842. IEEE, 2015.

\bibitem{ref4}	S. K. Ghosh, ``Visibility algorithms in the plane,'' \emph{Cambridge university press}, 2007.

\bibitem{ref5}	Lozano-P{\'e}rez, Tom{\'a}s and M. A. Wesley, ``An algorithm for planning collision-free paths among polyhedral obstacles,'' \emph{Communications of the ACM}, vol. 22, no. 10, pp. 560--570, 2007.

\bibitem{ref6}	F. Aurenhammer, ``Voronoi diagrams—a survey of a fundamental geometric data structure,'' \emph{ACM Computing Surveys (CSUR)}, vol. 23, no. 3, pp. 345--405, 1991.

\bibitem{ref7}	L. Kayraki,  P. Svestka,  J.C. Latombe, and M. Overmars, ``Probabilistic roadmaps for path planning in high-dimensional configurations spaces,'' \emph{Proc IEEE Trans Robot Autom}, vol. 12, no. 4, pp. 566--580, 1996.

\bibitem{ref8}	F. P. Preparata and M. I. Shamos, ``Computational geometry: an introduction,'' \emph{Springer-Verlag}, 1985. pp. 198--257.

\bibitem{ref9}	C. Goerzen Z. Kong, and B. Mettler. ``A survey of motion planning algorithms from the perspective of autonomous UAV guidance.'' \emph{Journal of Intelligent and Robotic Systems}, vol. 57, no. 1-4, pp. 65--100, 2010.

\bibitem{ref10} J. Van Den Berg, P. Abbeel, and K. Goldberg. ``LQG-MP: Optimized path planning for robots with motion uncertainty and imperfect state information.'' \emph{The International Journal of Robotics Research}, vol. 30, no. 7, pp. 895-913, 2011.

\bibitem{ref11} A. Ravankar, A. Ravankar, Y. Kobayashi, et al. ``Path smoothing techniques in robot navigation: State-of-the-art, current and future challenges,'' \emph{Sensors}, vol. 18, no. 9, 2018

\bibitem{ref12}N. Ratliff,  M. Zucker,  J. A. Bagnell, and S. Srinivasa,  ``CHOMP: Gradient optimization techniques for efficient motion planning,'' \emph{Robotics and Automation, 2009. ICRA'09. IEEE International Conference on}, pp. 489--494. IEEE, 2013.

\bibitem{ref13} J. Ulen, P. Strandmark, and F. Kahl. ``Shortest paths with higher-order regularization,'' \emph{IEEE transactions on pattern analysis and machine intelligence}, vol. 37, no. 12, pp. 2588-2600, 2015.

\bibitem{ref14} B. Song, G. Tian, F. Zhou. ``A comparison study on path smoothing algorithms for laser robot navigated mobile robot path planning in intelligent space,'' \emph{Journal of Information and Computational Science}, vol. 7, no. 1, pp. 2943-2950, 2010

\bibitem{ref15} L. Piegl, W. Tiller. ``The NURBS book,'' \emph{Springer Science \& Business Media}, 2012.

\bibitem{ref16} Geraerts, Roland and M. H. Overmars, ``Creating high-quality paths for motion planning,'' \emph{The International Journal of Robotics Research}, vol. 26, no. 8, pp. 845--863, 2007.

\bibitem{ref17} L. E. Dubins, ``On curves of minimal length with a constraint on average curvature, and with prescribed initial and terminal positions and tangents,'' \emph{American Journal of mathematics}, vol. 79, no. 3, pp. 497-516, 1957.

\bibitem{ref18} T. Fraichard, and A. Scheuer. ``From Reeds and Shepp's to continuous-curvature paths,'' \emph{IEEE Transactions on Robotics}, vol. 20, no. 6, pp. 1025-1035, 2004.

\bibitem{ref19} E. H. Lockwood, ``A book of curves,'' \emph{Cambridge University Press}, 1967.

\bibitem{ref20} C. Liu, C.-Y. Lin, and M. Tomizuka, ``The convex feasible set algorithm for real time optimization in motion planning,'' \emph{SIAM Journal on Control and Optimization}, vol. 56, no. 4, pp. 2712--2733, 2018.

\bibitem{ref21} C. Liu, C.-Y. Lin, Y. Wang, and M. Tomizuka, ``Convex feasible set algorithm for constrained trajectory smoothing,'' \emph{American Control Conference (ACC)}, pp. 4177--4182. IEEE, 2018.

\bibitem{ref22} C. Liu and M. Tomizuka, ``Real time trajectory optimization for nonlinear robotic systems: Relaxation and convexification,'' \emph{Systems \& Control Letters}, vol. 108, pp. 56--63. Elsevier, 2017.

\bibitem{ref23} C. Liu, ``Designing Robot Behavior in Human-Robot Interactions,'' Ph.D. dissertation, UC Berkeley, 2017.

\bibitem{ref24} E W Dijkstra. ``A note on two problems in connexion with graphs,'' \emph{Numerische mathematik}, vol 1, no. 1, pp. 269-271, 1959.

\bibitem{ref25} N Karmarkar. ``A new polynomial-time algorithm for linear programming,'' \emph{Proceedings of the sixteenth annual ACM symposium on Theory of computing. ACM}, pp. 302-311, 1984.

\bibitem{ref26} A. P. Teixeira, and R. Almeida. ``On the complexity of a mehrotra-type predictor-corrector algorithm,'' \emph{International Conference on Computational Science and Its Applications}. Springer, Berlin, Heidelberg, 2012.

\bibitem{ref27} A. Oktay, P. Tsiotras, and X. Huo. ``Solving shortest path problems with curvature constraints using beamlets,'' \emph{Intelligent Robots and Systems (IROS), 2011 IEEE/RSJ International Conference on}. IEEE, 2011.

\bibitem{ref28} D. L. Donoho, and X. Huo. ``Beamlets and multiscale image analysis,'' \emph{Multiscale and multiresolution methods}. Springer, Berlin, Heidelberg, pp. 149-196, 2002.

\bibitem{ref29} S. John. ``Filling space with random fractal non-overlapping simple shapes,'' \emph{Hyperseeing summer}, pp 131-140, 2011.

\bibitem{ref30} https://github.com/changliuliu/CFS

\bibitem{ref31} CVX http://cvxr.com/cvx/

\end{thebibliography}
\end{document}